\documentclass[10pt,twocolumn,letterpaper]{article}

\usepackage{cvpr}              

\usepackage{graphicx}
\usepackage{subcaption}
\usepackage{multirow} 
\newcommand{\cred}[1]{#1}

\newcommand{\sys}{\textsc{Jano}\xspace}

\definecolor{cvprblue}{rgb}{0.21,0.49,0.74}
\usepackage[pagebackref,breaklinks,colorlinks,allcolors=cvprblue]{hyperref}
\usepackage{xcolor}
\usepackage{multirow}
\usepackage{tabularx}
\usepackage{colortbl}
\usepackage[linesnumbered,noend]{algorithm2e}

\title{\sys: Adaptive Diffusion Generation with Early-stage Convergence Awareness}

\author{Yuyang Chen\thanks{Equal contribution.}\\
Shanghai Jiaotong University\\
{\tt\small chen-yy20@sjtu.edu.cn}
\and
Linqian Zeng\footnotemark[\value{footnote}]\\
Shanghai Jiaotong University\\
{\tt\small 028711zlq@sjtu.edu.cn}
\and
Yijin Zhou\\
Shanghai Jiaotong University\\
{\tt\small 708972751@sjtu.edu.cn}
\and
Hengjie Li\\
Shanghai AI Laboratory\\
{\tt\small lihengjie@pjlab.org.cn}
\and
Jidong Zhai\thanks{Corresponding author.}\\
Tsinghua University\\
{\tt\small zhaijidong@tsinghua.edu.cn}
}

\begin{document}
\maketitle

\begin{abstract}
Diffusion models have achieved remarkable success in generative AI, yet their computational efficiency remains a significant challenge, particularly for Diffusion Transformers (DiTs) requiring intensive full-attention computation. While existing acceleration approaches focus on content-agnostic uniform optimization strategies, we observe that different regions in generated content exhibit heterogeneous convergence patterns during the denoising process. We present \sys, a training-free framework that leverages this insight for efficient region-aware generation. \sys introduces an early-stage complexity recognition algorithm that accurately identifies regional convergence requirements within initial denoising steps, coupled with an adaptive token scheduling runtime that optimizes computational resource allocation. Through comprehensive evaluation on state-of-the-art models, \sys achieves substantial acceleration (average $2.0\times$ speedup, up to $2.4\times$) while preserving generation quality. Our work challenges conventional uniform processing assumptions and provides a practical solution for accelerating large-scale content generation. The source code of our implementation is available at \url{https://github.com/chen-yy20/Jano}.
\end{abstract}
\section{Introduction}

Recent advances in generative AI, particularly diffusion models, have revolutionized digital content creation with their unprecedented generation quality. As a notable milestone, Diffusion Transformers (DiTs)~\cite{dit} have gradually superseded traditional UNet~\cite{unet} architectures as the predominant paradigm for high-fidelity content generation, marking a significant advancement in model architectures.

Despite the impressive quality achievements, the computational efficiency of these models remains a critical bottleneck. For instance, generating merely a 5-second 720p video using the Wan2.1-14B~\cite{wan} model on a single NVIDIA A100 GPU requires over an hour for computation. This computational burden primarily stems from the full-attention mechanism in DiTs. Unlike LLMs' causal attention with KV Cache, DiTs require full-sequence attention across all tokens at every timestep, resulting in quadratic computational complexity with respect to sequence length. Such intensive computation poses significant challenges for real-time and interactive applications.

\begin{figure}[t]
    \centering
    \includegraphics[width=\linewidth]{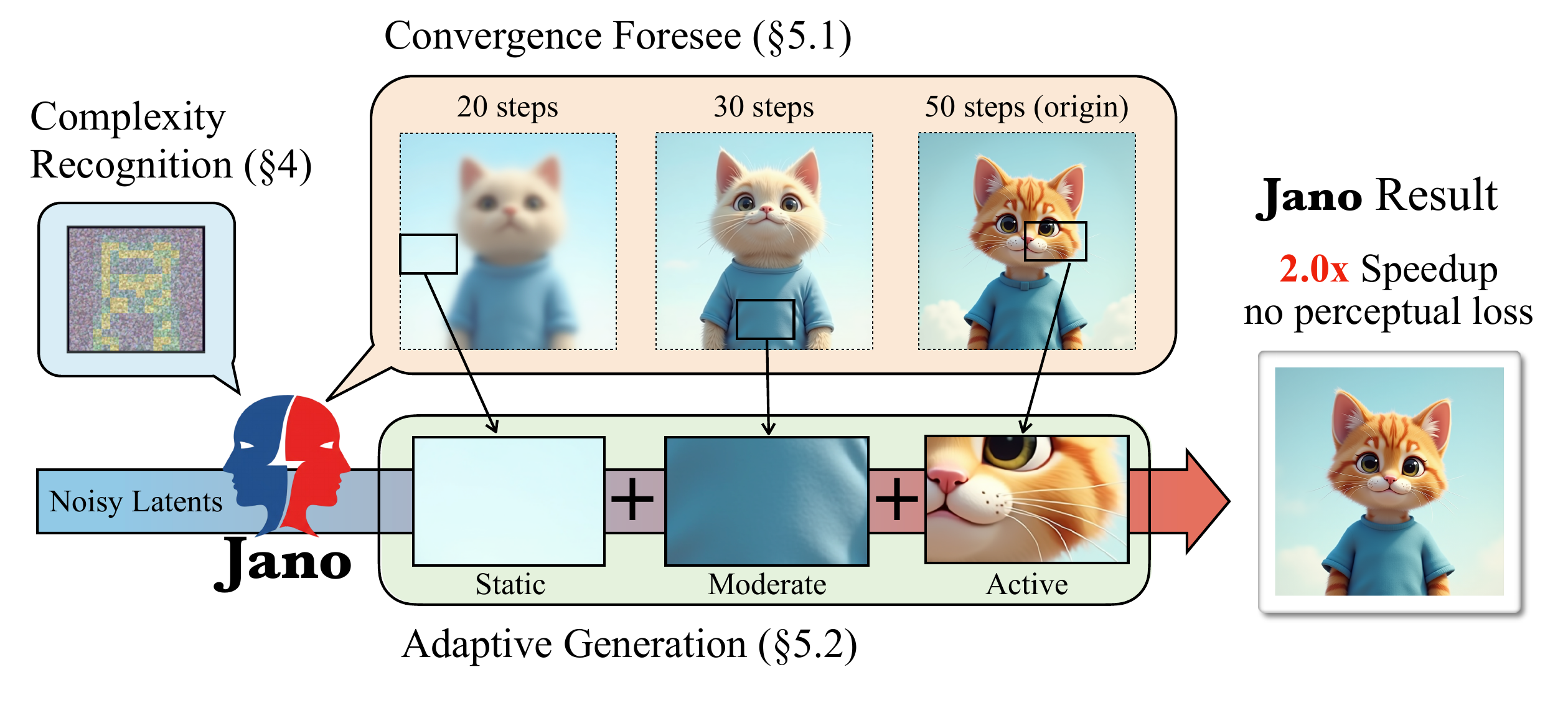}
    \caption{\sys foresees regional convergence through early-stage complexity recognition. Then it adaptively allocates computation resources, achieving $2.0\times$ speedup without perceptual loss. }
    \label{fig:overview}
\end{figure}

Existing train-free diffusion acceleration methods primarily rely on feature caching. However, these approaches face an inherent trade-off: coarse-grained, content-agnostic caching~\cite{adacache,teacache,magcache} fails to achieve optimal acceleration, while fine-grained caching at layer or token level~\cite{pab,tokencache,ras} requires costly per-step monitoring and management that introduces significant overhead.

We observe that different spatial regions of generated content exhibit distinct convergence behaviors — specifically, the number of denoising steps required for a latent value to stabilize varies significantly across regions. Furthermore, this convergence speed is strongly correlated with regional content complexity: semantically rich regions (e.g., facial features) demand sustained computational effort before reaching a stable state, whereas structurally simple regions (e.g., uniform backgrounds) converge within far fewer steps. This inherent non-uniformity in convergence dynamics reveals an opportunity for region-based computational optimization.

The key challenges in implementing this approach are twofold: accurately recognizing regional convergence patterns during early noisy steps, and efficiently orchestrating selective computation while maintaining compatibility with diffusion samplers and attention mechanisms that expect complete token sets.

We propose \sys, a training-free framework that enables early complexity recognition and adaptive generation optimization. Named after the Roman god who simultaneously observes past and future, \sys analyzes early-stage complexity patterns while foreseeing convergence trajectories. As illustrated in Fig.~\ref{fig:overview}, \sys introduces an algorithm that identifies regional complexity within early denoising steps, categorizing tokens into three convergence levels. This awareness enables convergence-adaptive generation through an efficient interleaved pipeline with KV Cache that selectively activates tokens based on their predicted convergence requirements.

The key contributions of this work are:
\begin{itemize}
\item We identify and characterize the heterogeneous nature of convergence patterns in diffusion generation and propose an approach to foresee regional convergence requirements during early noise-dominated stages.

\item We present \sys, a lightweight framework that integrates early-stage complexity recognition with an adaptive token activation strategy. Our system introduces minimal computational overhead through an efficient interleaved generation pipeline with optimized KV Cache management.

\item Through comprehensive evaluation on various generation tasks, we demonstrate that \sys achieves substantial acceleration average $2.0\times$ speedup, up to $2.4 \times$ on state-of-the-art diffusion models with agnostic quality loss.

\end{itemize}

\section{Related Work}

\subsection{Diffusion Models}
Diffusion models represent the state-of-the-art paradigm for AI-generated content, delivering exceptional quality across various media types. The generation process employs iterative denoising: starting from pure Gaussian noise and conditions (e.g. text prompts or images), the model progressively predicts and removes noise components across multiple steps until a clean sample emerges. While early implementations utilized UNet~\cite{unet} architectures, contemporary systems have largely transitioned to Diffusion Transformers (DiTs)~\cite{dit}, which offer superior scalability and generation capabilities. However, this architectural shift introduces significant computational challenges, as DiTs' self-attention mechanisms scale quadratically with sequence length, creating a critical performance bottleneck for high-resolution and long-duration content generation.

\subsection{Feature Caching Acceleration}
Feature caching has emerged as a simple yet effective approach to accelerate diffusion generation. By exploiting the temporal coherence between consecutive denoising steps, these methods reuse cached intermediate results to avoid redundant computations. Caching granularity varies across approaches, ranging from token-level~\cite{tokencache} and module-level~\cite{adacache, pab} to layer-level~\cite{deltadit, blockdance} and full transformer output-level~\cite{teacache, magcache}. However, indiscriminate reuse inevitably introduces generation quality degradation. To mitigate this, existing methods adopt different reuse strategies, including static schedules based on empirical observation~\cite{pab, deltadit}, step-wise reactive decisions~\cite{adacache, teacache, magcache}, and post-training optimization~\cite{deepcache, learningtocache}. Despite these efforts, these approaches apply uniform caching strategies regardless of content characteristics, leading to suboptimal quality-efficiency trade-offs.

\subsection{Region-Aware Diffusion}
Several works have explored region-aware processing in diffusion models, though with objectives distinct from \sys. Upsample What Matters~\cite{upsample} performs additional sampling on selected regions after low-resolution generation, while Region-Aware Diffusion Models~\cite{rad} and AdaDiffSR~\cite{adadiff} apply region-based processing for specific tasks such as inpainting and super-resolution. These methods operate on already-generated or partially-complete content, where region identification is relatively straightforward. Region-Aware Sampler~\cite{ras} monitors predicted noise variance to dynamically identify regions requiring recomputation during inference. However, its recognition accuracy remains insufficient, fundamentally constraining the achievable acceleration without compromising generation quality (see Appendix~\ref{sec:appendix_ras} for detailed analysis). In contrast, \sys addresses this challenge through early-stage complexity recognition, enabling more accurate and efficient region-aware acceleration.

\section{Motivation}
\label{sec:motivation}

\begin{figure}[t]
    \centering
    \includegraphics[width=\linewidth]{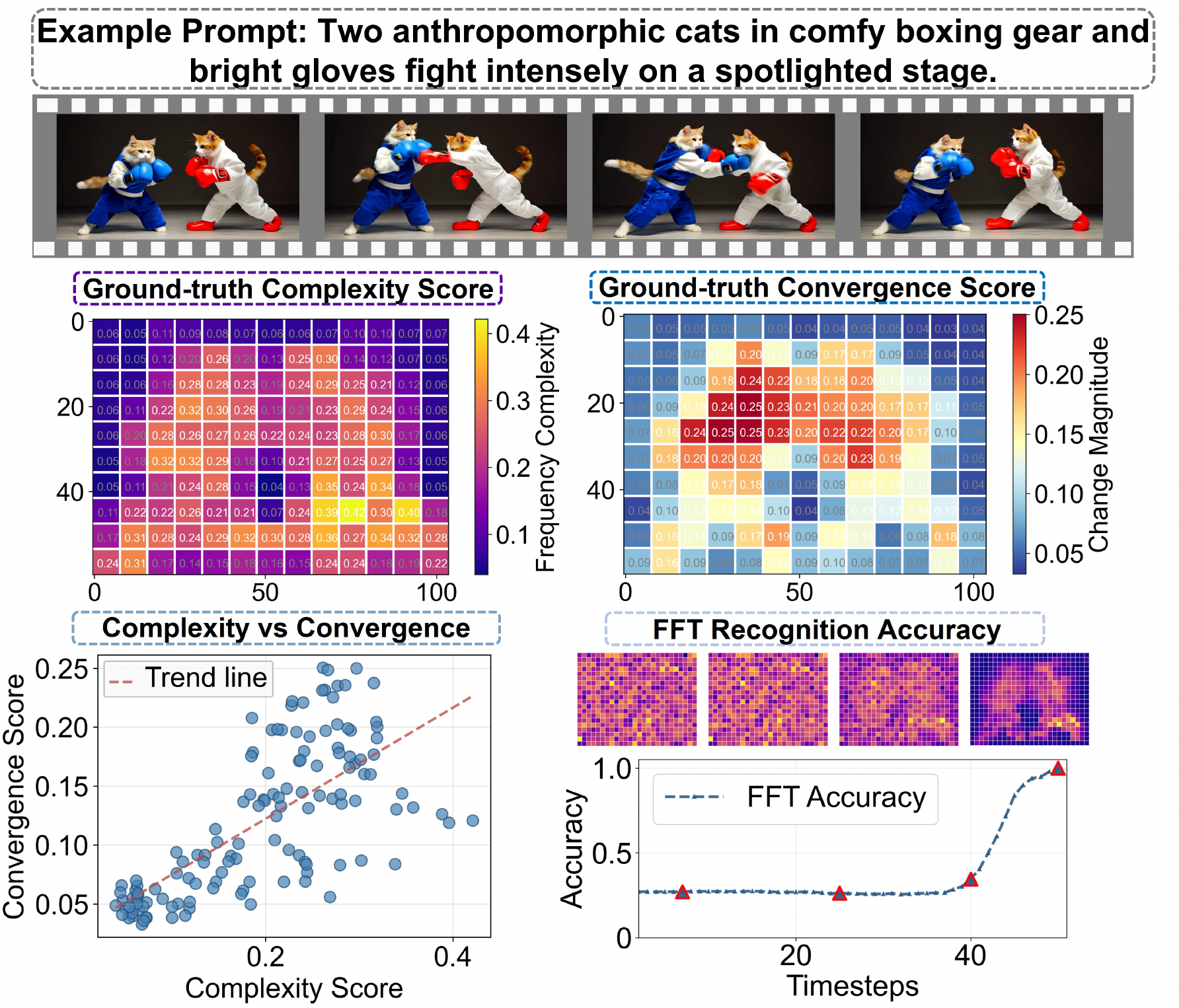}
    \caption{Motivation Example with complexity and convergence pattern analysis ($r=0.70,\ \rho=0.74$).
    }
    \label{fig:motiv}
\end{figure}

We present our key observation: different content regions exhibit distinct convergence behaviors that strongly correlate with their semantic complexity.

To validate this, we conduct a systematic analysis using Wan-1.3B~\cite{wan}. As shown in Fig.~\ref{fig:motiv}, we partition generated videos into fixed-size 3D blocks (frames$\times$height$\times$width) and analyze their complexity and convergence characteristics using two metrics:

\begin{itemize}
    \item \emph{Ground-Truth Complexity}: We measure the high-frequency energy distribution across channels and frames in the generated latent. For each 2D spatial slice, we apply a centered 2D FFT and calculate the energy ratio outside a quarter-radius circle in frequency space. The block's complexity is the average ratio across all channels and temporal dimensions.
    \item \emph{Ground-Truth Convergence}: We quantify how drastically the latent representation evolves during the denoising process. Specifically, starting from step 10, we compute the mean absolute difference between latent vectors at every 5-step interval. These differences are accumulated and normalized to obtain the ground-truth convergence, where a higher value indicates more substantial changes during denoising.
\end{itemize}

Fig.~\ref{fig:motiv} visualizes the block-wise complexity and convergence patterns of the example video, revealing a striking correlation. This observation is further validated across $50$ diverse video prompts in Fig.~\ref{fig:mapping} (a), consistently showing strong statistical significance (Spearman correlation $ \rho \approx 0.7, p < 0.001$).

The discovery of this non-uniform yet highly correlated complexity-convergence relationship suggests potential for content-aware optimization. However, leveraging this correlation during generation poses significant challenges. While our analysis reveals clear patterns in the final output, we need to foresee these patterns as early as possible during the generation process. As shown in the bottom-right panel of Fig.~\ref{fig:mapping}, using FFT-based complexity measurement on intermediate latents proves unreliable - when benchmarked against the complexity of fully denoised latents, the accuracy remains poor until late stages ($\rm{timestep}\approx 40$), only showing sharp improvement between $\rm{timestep}\in[40,50]$. This indicates that conventional approaches like FFT struggle to extract meaningful features from noise-dominated early-stage latents. This motivates our solutions in Section~\ref{sec:method} to recognize early-stage complexity and foresee the regional convergence.

\section{Early-stage Complexity Recognition}
\label{sec:method}
\subsection{Preliminary: Flow Matching}
\label{bg:flow-matching}

Traditional diffusion approaches follow a predefined noise schedule with discrete timesteps, where the generation process involves iterative denoising:

\begin{equation}
\mu(x_{t},t) = \frac{1}{\sqrt{\alpha_t}}(x_t - \frac{1-\alpha_t}{\sqrt{1-\bar{\alpha_t}}}\epsilon_\theta(x_t, t))
\end{equation}
where $\alpha_t$, $\bar{\alpha_t}$ are scheduling coefficients~\cite{DDPM,ddim}, and $\epsilon_\theta$ predicts the noise component to be removed.

Recent advances have introduced Flow Matching (FM)~\cite{flowmatching} as a more principled and flexible framework that reformulates the generation process as learning continuous normalizing flows. Unlike traditional diffusion models that operate through discrete denoising steps, FM directly learns a velocity field $v_\theta$ that parameterizes the continuous-time dynamics of the transformation from a simple prior distribution (typically Gaussian noise) to the target data distribution. 
A common instantiation, rectified flows \cite{liu2022rectifiedflowmarginalpreserving}, defines a linear interpolation between a source–target pair $(x_0, x_1)$ over continuous time $t\in[0,1]$:
\begin{equation}
x_t=tx_1+(1-t)x_0
\label{def}
\end{equation}
where $x_0 \sim \mathcal{N}(0,I)$ represents the initial noise and $x_1$ corresponds to samples from the target distribution. The velocity field $v_\theta(x_t, t)$ is trained to predict the instantaneous direction and magnitude of change at each point along these trajectories, enabling direct sampling through ODE integration.

FM is now widely used as the training paradigm in many state-of-the-art generative models.
Recent breakthroughs, including Flux~\cite{flux}, HunyuanVideo~\cite{hunyuan}, and Wan ~\cite{wan} have adopted FM as their core generation mechanism, leveraging DiTs~\cite{dit} architectures to predict velocity vectors $v_\theta: (x_t, t, c) \mapsto \mathbb{R}^d$ rather than traditional noise residuals. This approach offers improved sampling efficiency, better numerical stability, and more intuitive control over the generation dynamics compared to conventional diffusion formulations.

Based on flow-matching parameterization, we first derive our early-stage complexity recognition method. Through the $\varepsilon\!\to\!v$ transform, we further demonstrate in Appendix~\ref{app:fm-to-ddpm} that our framework generalizes naturally to other diffusion-based approaches, including DDIM~\citep{ddim}.

\subsection{Distance of latents}
\label{sec:theory}
To enable early-stage complexity recognition, we first develop a theoretical foundation for measuring content similarity between regions during generation. Based on the Flow Matching framework introduced in Section~\ref{bg:flow-matching}, the model $v_\theta$ is trained to regress the velocity field along the trajectory, following the linear path from $x_0$ to $x_1$, using the following training objective:

\begin{equation}
    \mathcal{L}_{FM} = \mathbb{E}_{t,\,x_0\sim \mathcal{N}(0,I),\,x_1\sim p(x_1)}    \big\| (x_1 - x_0) - v_\theta(x_t, t) \big\|^2.
\end{equation}

From regression theory, we know that the risk minimizer under squared loss corresponds to the conditional expectation:
\begin{equation}
\begin{aligned}
    v^\star_\theta(x_t,t) 
    &= \arg\min_{v_\theta}\;
       \mathbb{E}\!\left[\| (x_1-x_0) - v_\theta(x_t,t) \|^2 \right] \\
    &= \mathbb{E}\!\left[x_1 - x_0 \,\big|\, (x_t,t)\right].
\end{aligned}
\label{v_theta}
\end{equation}

Leveraging the interpolation relationship established in Eq. (\ref{def}), we can reformulate Eq. (\ref{v_theta}) as:
\begin{equation}
    v^\star_\theta(x_t,t) =\frac{\mathbb{E}\!\left[x_1\,\big|\, (x_t,t)\right]-x_t}{1-t}.
\end{equation}

To analyze the relationship between trajectories, we examine the velocity field difference between two points. Let $m(x_t,t)=\mathbb{E}\!\left[x_1\,\big|\, (x_t,t)\right]$ denote the conditional expectation of the target state. For any two trajectories A and B at time $t$, positioned at $x_{t,A}$ and $x_{t,B}$ respectively, applying Eq. (\ref{def}) yields: 
\begin{equation}
    v^\star_\theta(x_{t,B},t) - v^\star_\theta(x_{t,A},t) 
= (x_{0,B} - x_{0,A})+\Delta(x_{1,A},\;x_{1,B},\;t),
\end{equation}
where the correction term is given by $\Delta(x_{1,A},\;x_{1,B},\;t)=\frac{m_B - m_A}{1-t}
+ \frac{t}{1-t}(x_{1,B} - x_{1,A}).$ 

Applying the Lipschitz continuity property, we can bound this correction term:
\begin{equation}
   \Delta(x_{1,A},x_{1,B},t) \le \frac{t+f(t)}{1-t}\big\|x_{1,B}-x_{1,A}\big\|.
\end{equation}

A key insight emerges when considering trajectories that converge to similar outcomes. Specifically, when points A and B eventually reach similar final states, i.e., $x_{1,A}\approx x_{1,B}$, the difference in their velocities approaches a constant value across timesteps (Fig. \ref{fig:timestep}):
\begin{figure}[t]
    \centering
    \includegraphics[width=\linewidth]{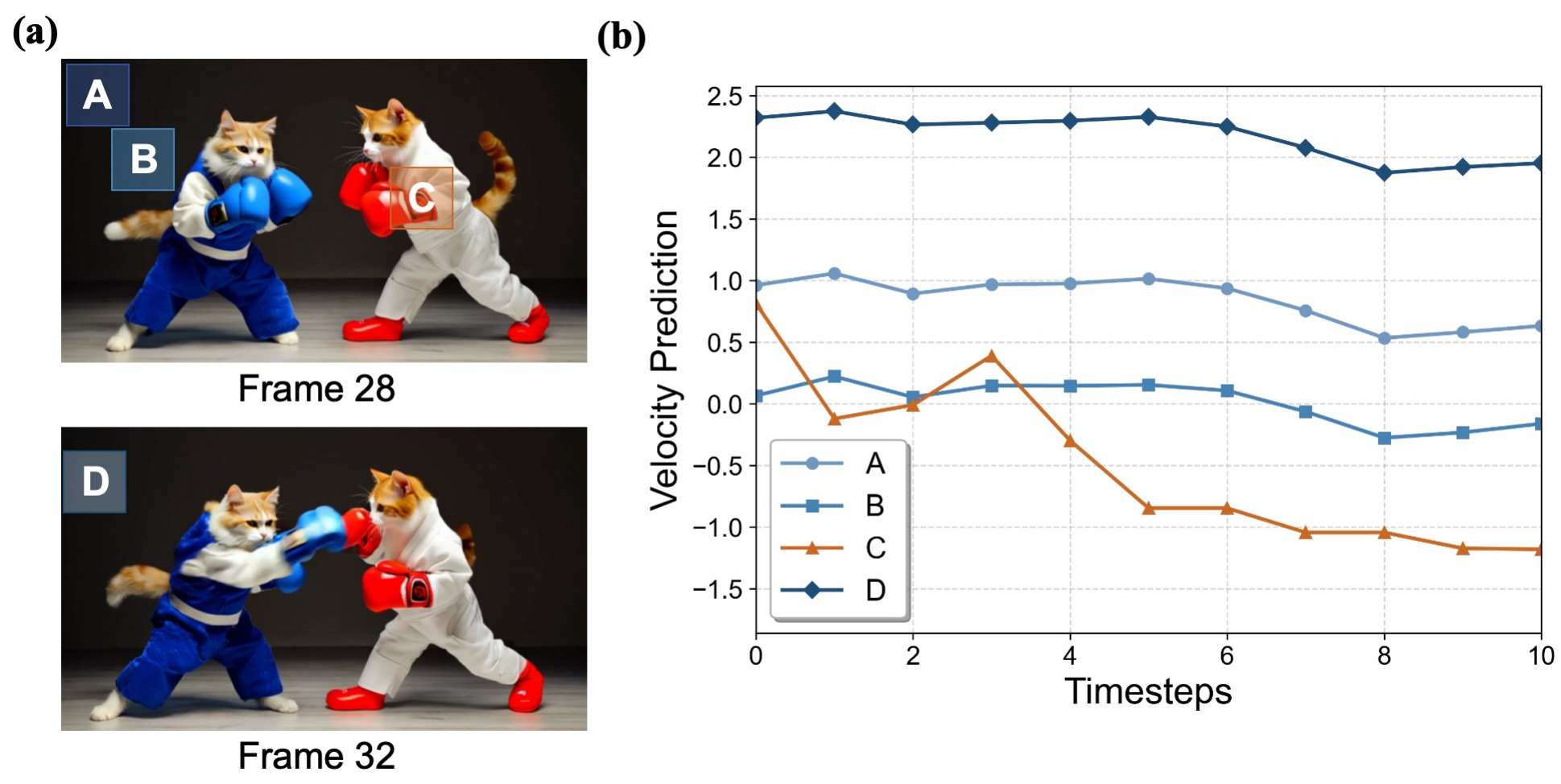}
    \caption{Velocity differences between both intra-frame similar points (A, B) and inter-frame similar points (A, D) remain approximately constant across timesteps.}
    \label{fig:timestep}
\end{figure}
\begin{align}
    \big\|v_\theta(x_{t,A},t) - v_\theta(x_{t,B},t)\big\| \approx \big\|x_{0,A} - x_{0,B}\big\|.
\end{align}

Building upon this theoretical foundation, we define a metric to quantify the predicted distance between latent representations A and B at timestep $t$:

\begin{equation}
    \label{eq:distance}
    D_{A,B}(t):=\big\|v_\theta(x_{t,A},t)-v_\theta(x_{t,B},t)\big\|-\big\|x_{0,A}-x_{0,B}\big\|.
\end{equation}

This distance metric $D_{A,B}(t)$ serves as an indicator of outcome similarity: larger values suggest that the trajectories will diverge to dissimilar final states. Importantly, this diagnostic capability emerges within the first few timesteps of the generation process, enabling early detection of whether any pair of points will converge to similar outcomes.

\subsection{Block-wise Complexity Analyzer}

We design a block-wise complexity analyzer inspired by Eq.~\eqref{eq:distance} for early-stage complexity recognition. While the distance metric relies on initial states $x_0$, we leverage interval-based block-wise differences across early steps to minimize error accumulation.

Given a latent tensor $\mathbf{x}\in\mathbb{R}^{C\times F\times H\times W}$ representing channel, frames, height, and width dimensions respectively, we first average it across channels and partition it into 3D-blocks of size $\cred{(f,h,w)}$.
For each block $b$ at diffusion step $k$, we obtain its block feature matrix by flattening spatial dimensions:
\begin{equation}
\mathbf{f}_b^{(k)}\in\mathbb{R}^{t\times s},\quad s=h{\cdot}w
\end{equation}

The complexity score incorporates both temporal and spatial dynamics through interval-based gradients. At timestep $k$, we compute the temporal gradient $\mathcal{T}_{b}^{(k)}$ to capture frame-wise variations:
\begin{equation}
\mathcal{T}_{b}^{(k)}
= \underset{s}{\mathrm{Average}}
\Bigl(\,
\Bigl\lVert
\sum_{(i,j)\in t}\bigl(f_{b}^{(k)}[i,s]-f_{b}^{(k)}[j,s]\bigr)
\Bigr\rVert_{2}
\Bigr)
\end{equation}
Similarly, the spatial gradient $\mathcal{S}_{b}^{(k)}$ measures local spatial changes:
\begin{equation}
\mathcal{S}_{b}^{(k)}
= \underset{t}{\mathrm{Average}}
\Bigl(\,
\Bigl\lVert
\sum_{(i,j)\in s}\bigl(f_{b}^{(k)}[t,i]-f_{b}^{(k)}[t,j]\bigr)
\Bigr\rVert_{2}
\Bigr)
\end{equation}

To analyze early-stage block-wise complexity, we examine latent tensors of $\cred{K}$ initial timesteps and compute second-order differences with interval $\Delta k=\lfloor \cred{K}/2\rfloor$, following Eq.~\eqref{eq:distance}:

\begin{align}
    \Delta \mathcal{T}_b &= \sum_{k=1}^{\lfloor \cred{K}/2 \rfloor} |\mathcal{T}_b^{(k+\Delta k)} - \mathcal{T}_b^{(k)}|, \\
    \Delta \mathcal{S}_b &= \sum_{k=1}^{\lfloor \cred{K}/2 \rfloor} |\mathcal{S}_b^{(k+\Delta k)} - \mathcal{S}_b^{(k)}|
\end{align}

The final complexity score combines temporal and spatial components with weighted summation:
\begin{equation}
\label{eq:complexity}
\mathcal{C}_b=\cred{\omega_t}\Delta \mathcal{T}_{b}+\cred{\omega_s}\Delta \mathcal{S}_{b},\quad\cred{\omega_t}+\cred{\omega_s}=1
\end{equation}
where weights $\cred{\omega_t}$ and $\cred{\omega_s}$ are chosen to emphasize temporal dynamics ($\cred{\omega_t} > \cred{\omega_s}$). For image generation ($F=1$), the score naturally simplifies to purely spatial gradients ($\mathcal{C}_b=\Delta \mathcal{S}_b$). Through empirical validation, we find that analyzing the first $\sim$\cred{10\%} of sampling steps provides an optimal balance between prediction reliability and computational efficiency, with the \cred{warm-up length} selected as the minimal step count where accuracy stabilizes (see Section~\ref{eval:accuracy}). Detailed parameter configurations are provided in Appendix~\ref{app:exp_params}.

\section{Convergence-Adaptive Generation}

\subsection{Convergence Foresee}

\begin{figure}[t]
    \centering
    \includegraphics[width=\linewidth]{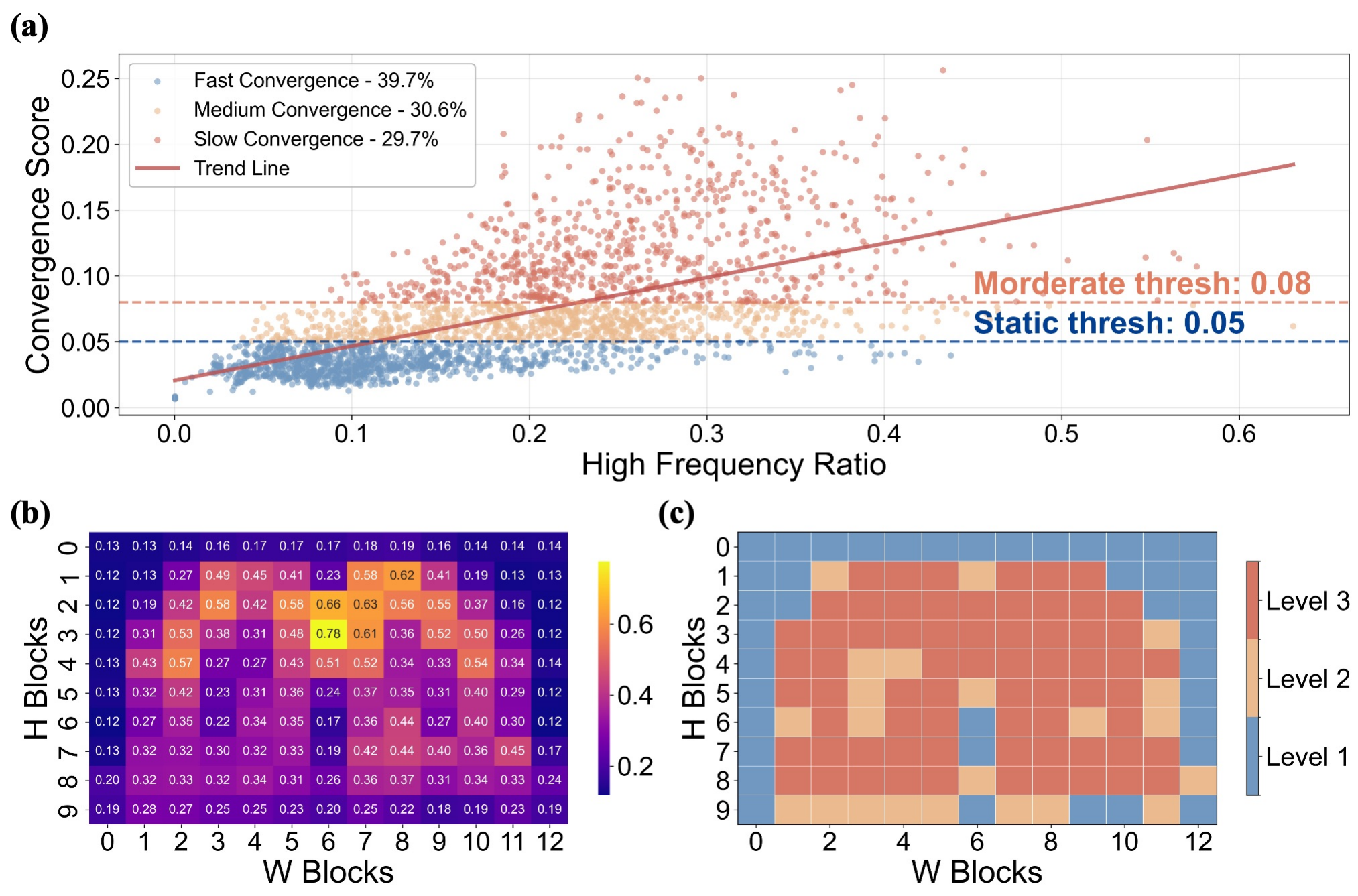}
    \caption{Statistical validation and visualization of complexity-driven convergence-level categorization. (a) Convergence score exhibits significant correlation with complexity across the dataset ($r = 0.61, \ \rho = 0.69,\ p < 0.001$), naturally forming three distinct levels. Predicted complexity scores (b) are mapped to three convergence levels (c) through optimized thresholds.}
    \label{fig:mapping}
    \vspace{-0.5\intextsep}
\end{figure}

To leverage the dynamic convergence patterns for generation acceleration, we introduce a three-level convergence classification based on correlation analysis in Fig.~\ref{fig:mapping}(a). This classification system categorizes all tokens according to their corresponding region's convergence behavior. Taking a multi-step generation process as an example, tokens are assigned to activity levels as follows:
\begin{itemize}
\item Level 1 (Static): Tokens in regions that converge within the early phase of denoising
\item Level 2 (Moderate): Tokens in regions requiring an intermediate number of steps
\item Level 3 (Active): Tokens in regions needing the full denoising process for completion
\end{itemize}

Statistically, across diverse generation tasks, these three levels maintain a balanced distribution that varies based on content complexity. For individual generation instances, the actual proportions depend on the relative complexity of different content regions within the specific sample being generated. This tripartite classification enables adaptive computational allocation by matching processing intensity to the inherent complexity requirements of different spatial regions.

As illustrated in Fig.\ref{fig:mapping}(b) and Fig.\ref{fig:mapping}(c), the early-stage recognized complexity scores are mapped to these convergence levels using optimized thresholds. These thresholds are derived through a heuristic algorithm that maximizes active block coverage while minimizing the overall computational cost (see Appendix~\ref{app:exp_params} for detailed parameter settings).

\subsection{Interleaved Generation with KV Cache}
\label{sec:pipeline}

\begin{figure}
    \centering
    \includegraphics[width=\linewidth]{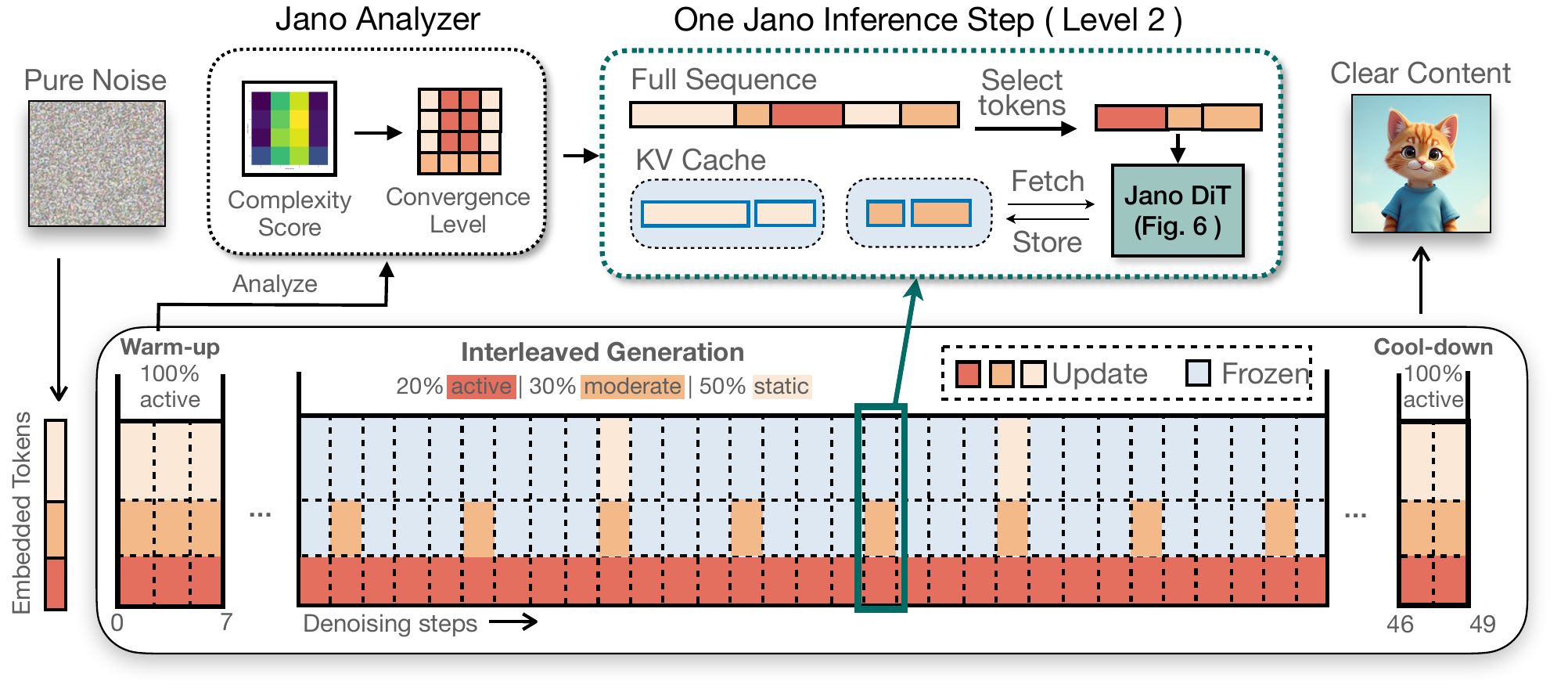}
    \caption{\sys adaptively computes tokens of different convergence levels through an interleaved pipeline}
    \label{fig:multi-level-generation}
\vspace{-0.5\intextsep}
\end{figure}

\paragraph{Interleaved Multi-level Pipeline} 
We design a three-phase generation pipeline that adaptively allocates computational resources based on regional convergence requirements (Fig.~\ref{fig:multi-level-generation}).

The \textbf{warm-up phase} analyzes early-stage outputs to establish initial complexity patterns. During this phase, we store and process output latents to construct a convergence level map, which guides subsequent resource allocation. This early-stage analysis is crucial for accurate complexity detection while minimizing computational overhead.

In the \textbf{interleaved generation phase}, blocks receive differentiated \cred{update frequencies} based on their convergence levels. Level-1 (static) blocks are activated at the lowest frequency, Level-2 (moderate) blocks at an intermediate frequency, while Level-3 (active) blocks remain continuously updated throughout generation. Note that Fig.~\ref{fig:multi-level-generation} presents one illustrative configuration; the specific \cred{update frequencies} are adjustable to achieve different quality--speed trade-offs (see Appendix~\ref{app:exp_params}).

The pipeline concludes with a \textbf{cool-down phase} focused on final refinement and detail enhancement. This phase ensures smooth transitions between regions of different convergence levels and refines local details, particularly in areas where sparse updates might have introduced minor artifacts. This final stage is essential for maintaining global coherence and achieving high-quality outputs.

\begin{figure}
    \centering
    \includegraphics[width=0.8\linewidth]{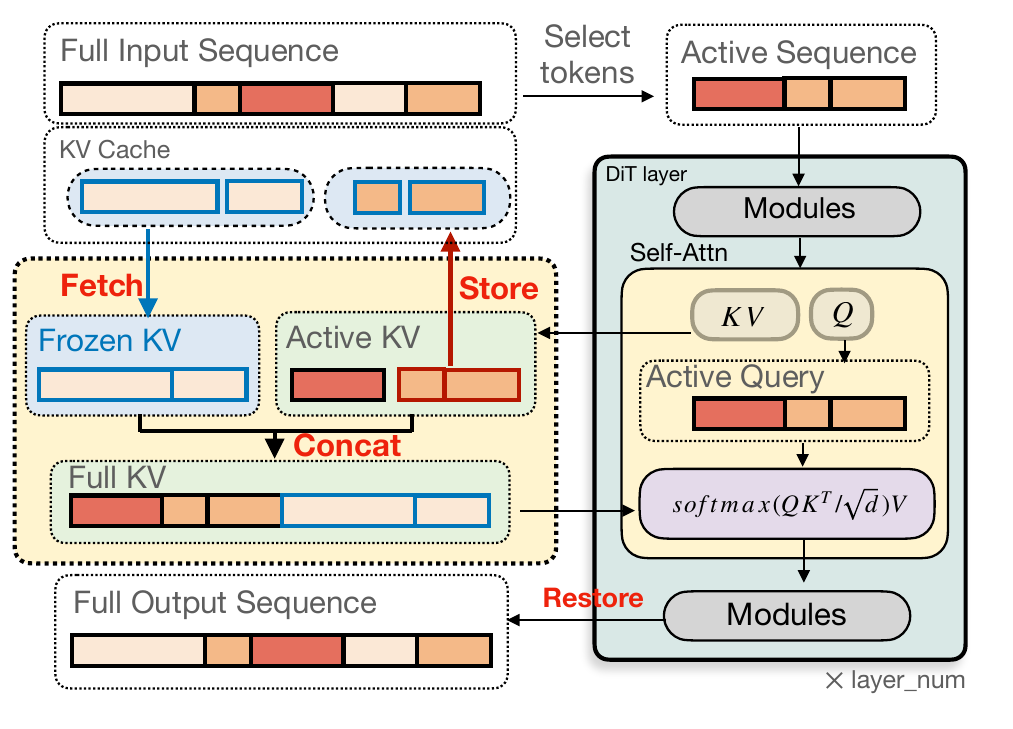}
    \caption{\sys DiT with KV Cache. Cached Key-Value pairs are fetched and concatenates to maintain global sequence dependency in full attention.}
    \label{fig:masked-dit}
    \vspace{-0.5\intextsep}
\end{figure}

\paragraph{KV Cache for DiT} 
The interleaved computation pattern presents dependency challenges in full attention, which necessitates global sequence information. To maintain computational consistency while only processing active tokens, we introduce a diffusion-specific KV cache mechanism. As shown in Fig.~\ref{fig:masked-dit}, the cache stores Key-Value pairs (KVs) of frozen tokens per convergence level. During token freezing periods, cached KVs are fetched and concatenated to complete the full KV sequence for attention computation with active query sequences. The cache updates when tokens are reactivated, ensuring efficient yet accurate attention mechanisms across varying activation patterns.

A key optimization in our KV retrieval strategy significantly reduces computational overhead. Instead of restoring cached KVs to their original positional indices, we directly concatenate all active and cached Key-Value pairs in sequence. This concatenation-based approach preserves the correctness of attention computation since attention mechanisms are inherently permutation-invariant with respect to key-value ordering, while dramatically reducing the overhead of cache operations. The direct concatenation eliminates costly positional restoration steps, making KV cache storage and retrieval operations virtually overhead-free.

\section{Evaluation}

\subsection{Setup}
\paragraph{Models and Baselines} We evaluate \sys on Flux-1-dev~\cite{flux} for image generation and Wan2.1-T2V (1.3B and 14B variants)~~\cite{wan} for video generation. Table~\ref{tab:model_specs} summarizes their specifications. We compare against state-of-the-art feature caching methods: TokenCache (ToCa)~~\cite{tokencache}, Pyramid Attention Broadcast (PAB)~~\cite{pab}, and TeaCache~\cite{teacache}. Evaluations use the t2i-diversity-eval~\cite{brack2025howtotrain} and VBench~\cite{vbench} datasets for images and videos respectively.

\begin{table}[h]
    \centering
    \footnotesize
    \caption{Evaluated Model Specifications}
    \begin{tabular}{c|c|c|c}
        \hline
        \textbf{Model} &\textbf{Resolution} & \textbf{Latent} & \textbf{Seq Length} \\
        \hline
        Flux-1 & 1024x1024 & [16,1,128,128] & 4096 \\
        Wan-1.3B & 832×480  & [16,21,60,104] & 32760 \\
        Wan-14B & 720×1280  & [16,21,90,160] & 75600 \\
        \hline
    \end{tabular}
    \label{tab:model_specs}
    \vspace{-\intextsep}
\end{table}

\paragraph{Metrics} We measure acceleration using end-to-end latency of 50-step generation. Quality assessment employs PSNR, SSIM, and LPIPS for both images and videos, with additional VBench metrics (ImageQual and SubConsist) for video evaluation.

\paragraph{Implementation} \sys is implemented in PyTorch 2.5. We implement compatible versions of ToCa for Wan and PAB for both Flux and Wan, as no official implementations are available, and use the official implementation of TeaCache.  Experiments are conducted on NVIDIA H800 80GB GPU, except for Wan-14B which uses two H200 141GB GPUs with conditional-free guidance parallelism~\cite{cfg, xdit}. Detailed parameter configurations for each model are listed in Appendix~\ref{app:exp_params}.

\subsection{Quality \& Efficiency}
\begin{figure*}[!ht]
    \centering
\includegraphics[width=0.9\linewidth]{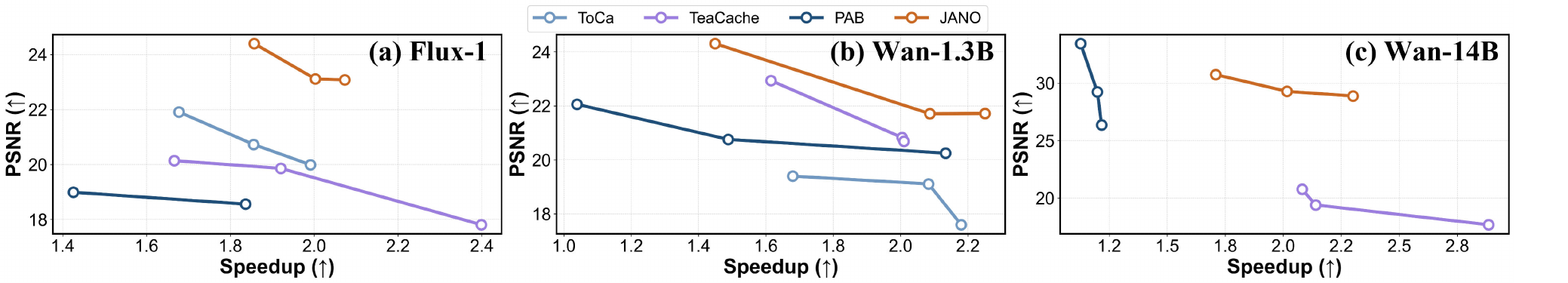}
    \caption{\sys advanced the Pareto frontier in the quality-latency tradeoff, achieving a better balance between generation quality and speed.
    }
    \label{fig_tradeoff}
\end{figure*}
\begin{table*}[!ht]
    \centering
    \caption{Quantitative evaluation of inference efficiency and visual quality in generation models.}
    \small
    \setlength{\tabcolsep}{15pt}
    \resizebox{\linewidth}{!}{%
    \renewcommand{\arraystretch}{0.9}
    \begin{tabular}{c|cc|ccccc}
    \toprule
        \multirow{2}{*}{\textbf{Method}} & \multicolumn{2}{c|}{\textbf{Efficiency}} & \multicolumn{5}{c}{\textbf{Visual Quality}} \\ \cmidrule(lr){2-8}
        ~ & Speedup $\uparrow$ & Latency (s) $\downarrow$ & LPIPS $\downarrow$ & SSIM $\uparrow$ & PSNR $\uparrow$ & ImageQual$\uparrow$ & SubConsist$\uparrow$ \\ \midrule
        \textbf{Flux-1 $(T=50)$} & 1$\times$ & 11.37 & - & - & - & - & - \\ \midrule
        PAB & 1.43$\times$ & 7.98 & 0.2866 & 0.7141  & 18.99  & - & -\\ 
        TeaCache & 1.67$\times$ & 6.83 & 0.1770 & 0.8014  & 20.14  & - & \\ 
        ToCa & 1.68$\times$ & 6.78 & 0.1487 & 0.8223 & 21.91 & - & - \\ 
        \rowcolor{gray!20}
         \sys & \textbf{1.86}$\times$ & \textbf{6.12} & \textbf{0.1310} & \textbf{0.8533} & \textbf{24.40} & - & - \\ 
         \midrule
        \textbf{Wan-1.3B $(T=50)$} & 1$\times$ & 103.3 & - & - & - & 0.7082 & 0.9704 \\ \midrule
        PAB & 1.49$\times$ & 69.42 & 0.1092 & 0.8188 & 20.76 & 0.7037 & 0.9691 \\ 
        TeaCache & 2.00$\times$ & 51.53 & 0.1056 & 0.8290 & 20.83 & 0.7031 & 0.9671 \\ 
        ToCa & 2.18$\times$ & 47.38 & 0.1868 & 0.7309 & 17.60 & 0.7002 & 0.9653 \\ 
        \rowcolor{gray!20}
        \sys & \textbf{2.25}$\times$ & \textbf{45.14} & \textbf{0.1033} & \textbf{0.8305} & \textbf{21.72} & \textbf{0.7051} & \textbf{0.9728} \\ 
        \midrule
        \textbf{Wan-14B $(T=50)$} & 1$\times$ & 1790.1 & - & - & - & 0.6839 & 0.9837 \\ \midrule
        PAB  & 1.22$\times$ & 1470.4 & 0.0785 & 0.8048 & 26.36 & 0.6736 & 0.9823 \\ 
        TeaCache & 2.08$\times$ & 860.3 & 0.2077 & 0.7420 & 19.39 & \textbf{0.6817} & 0.9817 \\ 
        \rowcolor{gray!20}
        \sys & \textbf{2.30}$\times$ & \textbf{778.4} & \textbf{0.0737} & \textbf{0.8658} & \textbf{28.87} & 0.6650 & \textbf{0.9862} \\
        \bottomrule
    \end{tabular}
    }
    \label{tab:main}
    \vspace{-0.5\intextsep}
\end{table*}

\paragraph{Quantitative Comparison}

We evaluate the quality and efficiency of images and videos generated with \sys against baseline methods, and summarize the results in Table~\ref{tab:main}. Notably, when applied to the Wan-14B model, ToCa exceeds the available GPU memory and leads to out-of-memory failures. The results show that \sys attains higher speedup while maintaining, or even improving, generation quality. 

Notably, because \sys reduces computation primarily in visually simple regions (e.g., backgrounds), it tends to attenuate background details while making the main subject more salient. As a result, subject consistency can even surpass that of the original video, although the overall image quality of individual frames may decrease slightly due to the weakened background. In terms of efficiency, since video generation introduces an additional temporal dimension, \sys achieves more pronounced acceleration on text-to-video models than text-to-image models. Moreover, as the number of parameters in the generative model increases, transformer computation dominates a larger fraction of the total generation time, further amplifying the advantage of \sys in delivering higher speedup without sacrificing quality.

\paragraph{Quality-Latency Trade-off}

In Fig.~\ref{fig_tradeoff}, we compare the PSNR–speedup trade-off of \sys against other baselines across different models. Over a broad range of speedup ratios, \sys consistently achieves higher PSNR, demonstrating a clear advantage in generation quality.

\subsection{Complexity Recognition Accuracy}
\label{eval:accuracy}

\begin{figure}[b]
    \centering
    \includegraphics[width=\linewidth]{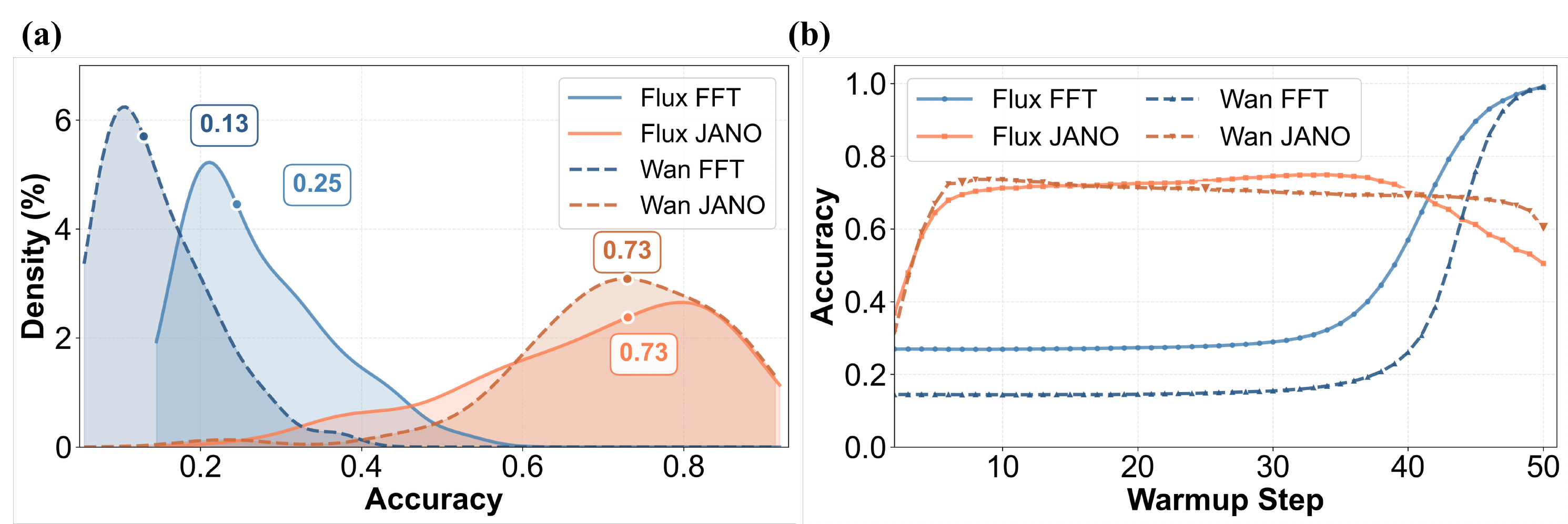}
    \caption{Early-stage complexity recognition on Flux and Wan 2.1.
(a) Prompt-averaged accuracy versus warm-up step. 
(b) Accuracy distributions across prompts at early-stage (Flux: 6 steps; Wan: 7 steps).}
    \label{fig_complexity}
\end{figure}

Fig.~\ref{fig_complexity}~(a) reports the mean accuracy of complexity identification over multiple prompts under different \cred{warm-up steps}. Using the FFT computed on the fully denoised latent as the reference, we observe that for Wan2.1 and Flux with \cred{warm-up of 6 and 7 steps}, respectively, \sys already achieves high accuracy, far exceeding FFT computed directly on latents from warm-up timesteps. Extending to a suite of prompts, when the \cred{warm-up is set to 6 steps for Wan 2.1 and 7 steps for Flux}, the median accuracy of \sys reaches 0.73—substantially higher than the FFT baselines on early-timestep latents (0.13 and 0.25). Overall, compared with FFT, \sys markedly improves complexity identification at early timesteps (See Fig.~\ref{fig_complexity}~(b)).

\begin{figure*}[h!]
    \centering
    \includegraphics[width=\textwidth]{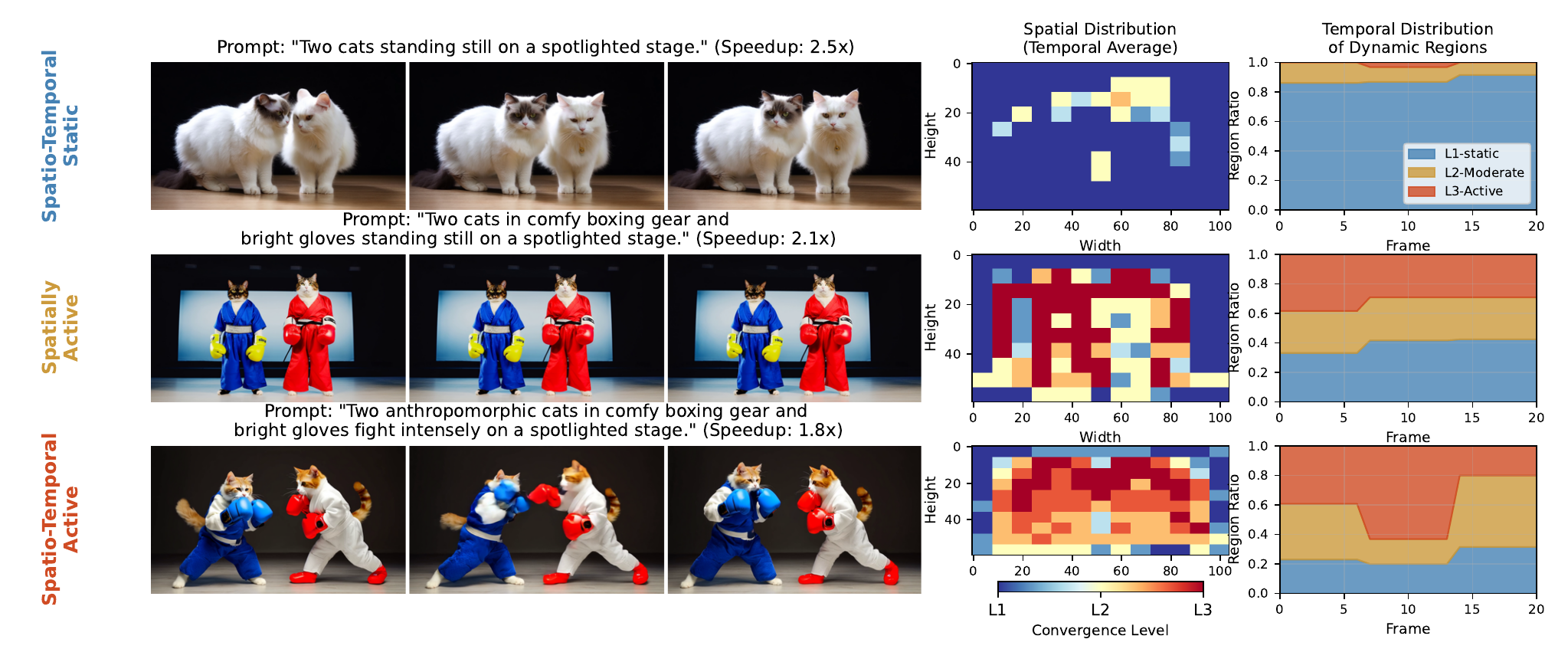}
    \caption{\sys sensitivity analysis across different dynamic scenarios.}
    \label{fig:sensitivity-test}
    \vspace{-0.5\intextsep}
\end{figure*}

\subsection{Sensitivity \& Breakdown Analysis}
\paragraph{Adaptive acceleration with different complexity.} \sys demonstrates effective content-aware resource allocation through early-stage convergence analysis across scenarios of varying spatio-temporal complexity. As shown in Fig.~\ref{fig:sensitivity-test}. It accurately identifies content dynamics: achieving 2.5× speedup for static scenes (stationary cats), 2.1× for spatially active content (cats in boxing gear), and 1.8× for spatio-temporally active scenes (fighting cats), where high convergence weights precisely track the motion-intensive regions. This adaptive identification enables an average speedup of 2.2× while preserving generation quality across diverse content complexities.

\begin{figure}
    \centering
    \includegraphics[width=\linewidth]{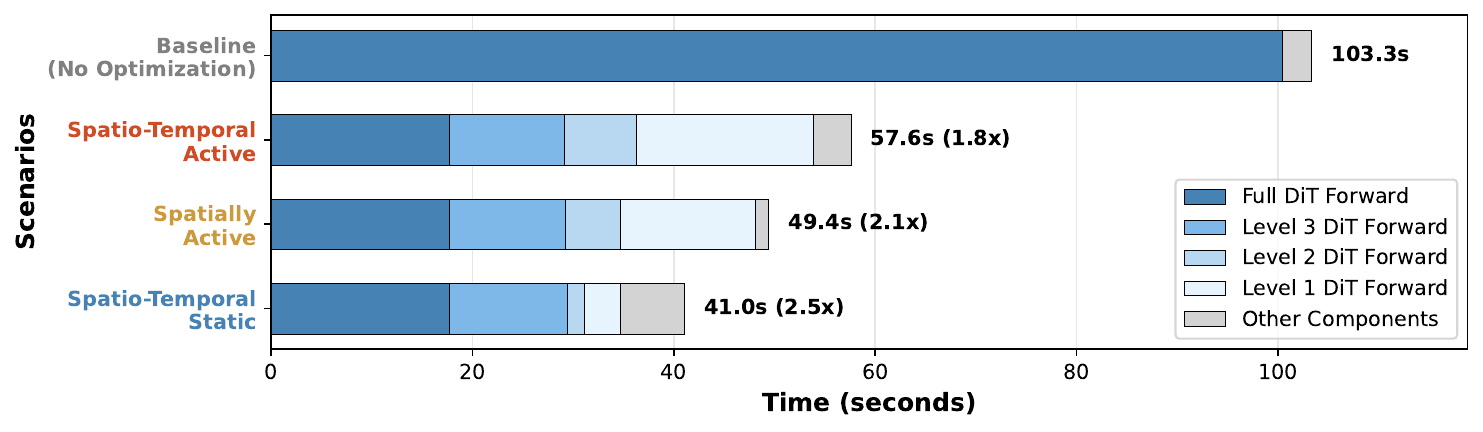}
    \caption{Detailed time breakdown of \sys generation under different token complexity scenarios.}
    \label{eval:breakdown}
    \vspace{-0.5\intextsep}
\end{figure}

\paragraph{Interleaved Pipeline Generation Breakdown} Fig.~\ref{eval:breakdown} illustrates the time distribution across different phases of \sys's generation pipeline under varying complexity scenarios. During the interleaved generation, Full DiT forward operations in warmup and cooldown phases, which process complete sequences without KV cache; Level 3 DiT operations that maintain constant computation time due to full sequence processing. Adaptive Level 1 and 2 DiT computations that handle moderate and static tokens respectively. The performance gains primarily stem from this adaptive computation strategy - as the proportion of moderate and static tokens varies across scenarios, the computational costs of Level 1 and 2 operations scale accordingly, enabling efficient resource utilization while maintaining generation quality.

\subsection{Ablation Study}
\begin{figure}
    \centering
    \includegraphics[width=\linewidth]{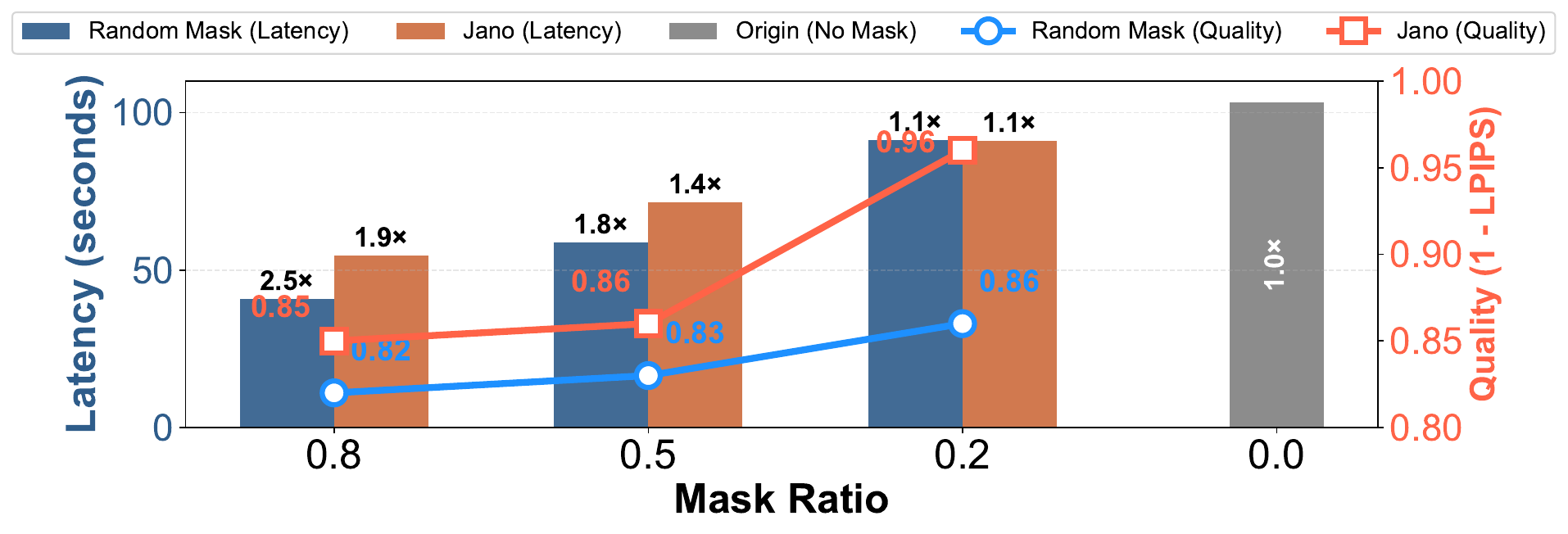}
    \caption{Ablation study comparing random masking versus \sys's convergence-aware masking}
    \label{eval:ablation}
    \vspace{-0.5\intextsep}
\end{figure}

To validate the effectiveness of convergence awareness in \sys, we compare it against random masking under different mask ratios, where masked latents are treated as static tokens in the interleaved pipeline. As shown in Fig.~\ref{eval:ablation}z, while both approaches achieve acceleration through the interleaved pipeline, \sys exhibits slightly lower speedup due to its handling of moderate tokens. However, \sys maintains significantly better generation quality and rapidly approaches optimal quality as mask ratio decreases. This demonstrates that our interleaved pipeline enables near-linear speedup, with early-stage convergence awareness playing a crucial role in preserving generation quality.

\section{Conclusion}

In this study, we theoretically characterize the relationship between the diffusion model’s denoised output and the final latent distance, and leverage this insight to propose \sys. This novel training-free method identifies image and video complexity in the early stages of generation and achieves DiT acceleration at the token level. Across representative state-of-the-art generative models, \sys delivers substantial end-to-end speedups while preserving output quality.

{
    \small
    \bibliographystyle{ieeenat_fullname}
    \bibliography{main}
}

\clearpage
\setcounter{page}{1}
\appendix
\maketitlesupplementary

\section{Applicability of \sys in DDIM}
\label{app:fm-to-ddpm}

We further investigate the complexity recognition algorithm of \sys to validate its effectiveness across different sampling methods, including non-flow-matching samplers such as DDIM~\cite{ddim}.

\begin{figure}[t]
  \centering
  \includegraphics[width=\linewidth]{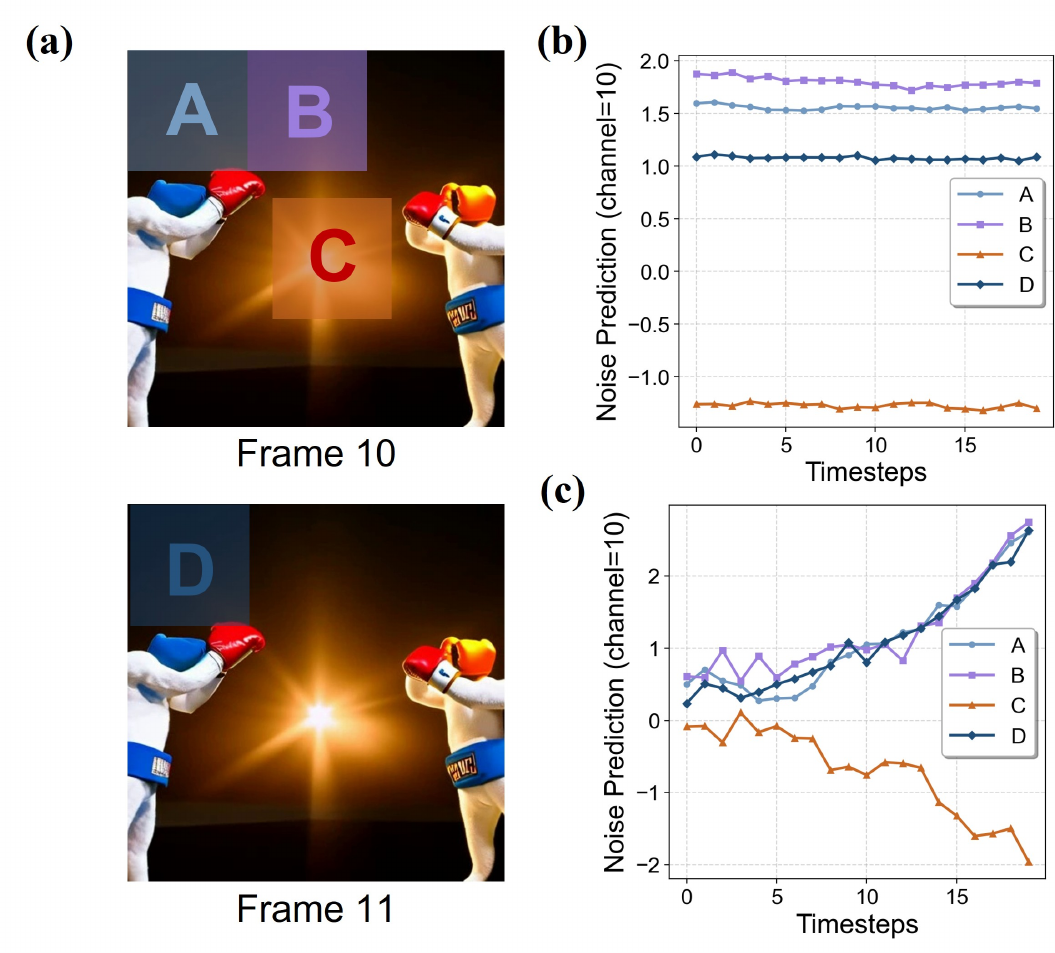}
  \caption{Latte video generation. (a) Frames 10 and 11 from the Latte-generated clip; markers A--D indicate four probe locations---the same A--D indices used in panels (b) and (c). (b) Raw DDIM model output at A--D across reverse-diffusion timesteps, before any transformation. (c) The same signals after applying the epsilon-to-FM linear mapping (Eq.~(\ref{transfer})).}
  \label{appendix_1}
\end{figure}

In DDIM, the denoiser is trained in the epsilon–prediction regime under the forward path
\[
x_t=\alpha(t)\,x_{\text{data}}+\sigma(t)\,\varepsilon,\qquad
\alpha(t)=\sqrt{\bar\alpha_t},\ \ \sigma(t)=\sqrt{1-\bar\alpha_t},
\]
and the network output satisfies the MMSE identity
\[
\hat\varepsilon_\theta(x_t,t)=\mathbb{E}[\varepsilon\,\big|\, x_t].
\]
Flow matching targets the conditional mean instantaneous velocity along the same path,
\[
v^*(x_t,t)=\mathbb{E}[\dot x_t\,\big|\, x_t],\qquad
\dot x_t=\dot\alpha(t)\,x_{\text{data}}+\dot\sigma(t)\,\varepsilon .
\]
Using
\[
x_t=\alpha(t)\,\mathbb{E}[x_{\text{data}}\,\big|\,x_t]+\sigma(t)\,\mathbb{E}[\varepsilon\,\big|\, x_t],\]
where $\mathbb{E}[x_{\text{data}}\,\big|\, x_t]=\frac{x_t-\sigma(t)\,\hat\varepsilon_\theta(x_t,t)}{\alpha(t)}$.
The DDIM epsilon output can be deterministically converted to the flow–matching velocity field:
\begin{equation}
\hat v_{\mathrm{FM}}(x_t,t)
=\frac{\dot\alpha(t)}{\alpha(t)}\,x_t
+\Bigg(\dot\sigma(t)-\frac{\dot\alpha(t)\,\sigma(t)}{\alpha(t)}\Bigg)\hat\varepsilon_\theta(x_t,t)
.\label{transfer}
\end{equation}
In discrete schedulers, the derivatives are obtained by finite differences between adjacent indices,
\[
\dot\alpha(t)\approx\frac{\alpha(t)-\alpha(t-\Delta t)}{\Delta t},\qquad
\dot\sigma(t)\approx\frac{\sigma(t)-\sigma(t-\Delta t)}{\Delta t},
\]
After applying this transformation, the resulting estimates demonstrate properties consistent with flow matching theory. We validate this empirically using the DDIM sampler on Latte~\cite{latte}, as illustrated in Fig.~\ref{appendix_1}.

\section{Experimental Parameters}
\label{app:exp_params}

For the results reported in the main text, JANO is configured with the parameter settings listed in Table~\ref{tab:model_specs}, and the warmup steps are the same as those used in Section~6.3. To separately assess image and video generation quality, we conduct experiments on prompts from the t2i-diversity-evalprompts dataset \cite{brack2025howtotrain} for image generation and from the VBench benchmark~\cite{vbench} for video generation.
\begin{table*}[h]
    \centering
    \caption{\sys parameter settings and performance.}
    \label{tab:exp_params}
    {
    \begin{tabular}{l c c c c c c c c}
        \toprule
        \multirow{2}{*}{Model}
        & \multicolumn{2}{c}{Static}
        & \multicolumn{2}{c}{Moderate}
        & \multirow{2}{*}{Speedup~$\uparrow$}
        & \multirow{2}{*}{LPIPS~$\downarrow$}
        & \multirow{2}{*}{SSIM~$\uparrow$}
        & \multirow{2}{*}{PSNR~$\uparrow$} \\
        \cmidrule(r){2-3}\cmidrule(lr){4-5}
        & \cred{threshold} & \cred{interval} & \cred{threshold} & \cred{interval} & & & & \\
        \midrule
        Flux-1
        & \cred{0.1}  & \cred{8}  & \cred{0.5} & \cred{5} & 1.86 & 0.1310 & 0.8533 & 24.40 \\
        & \cred{0.2}  & \cred{10} & \cred{0.5} & \cred{3} & 2.00 & 0.1728 & 0.8182 & 23.11 \\
        & \cred{0.3}  & \cred{10} & \cred{0.6} & \cred{3} & 2.07 & 0.1768 & 0.8153 & 23.08 \\
        \midrule
        Wan-1.3B
        & \cred{0.1}   & \cred{3}  & \cred{0.3} & \cred{2} & 1.45 & 0.0588 & 0.8799 & 24.31 \\
        & \cred{0.15}  & \cred{6}  & \cred{0.4} & \cred{4} & 2.09 & 0.1001 & 0.8325 & 21.71 \\
        & \cred{0.4}   & \cred{6}  & \cred{0.6} & \cred{4} & 2.25 & 0.1033 & 0.8305 & 21.71 \\
        \midrule
        Wan-14B
        & \cred{0.1}   & \cred{6}  & \cred{0.3} & \cred{3} & 1.71 & 0.0640 & 0.8723 & 30.73 \\
        & \cred{0.15}  & \cred{6}  & \cred{0.4} & \cred{4} & 2.02 & 0.0732 & 0.8646 & 29.27 \\
        & \cred{0.4}   & \cred{6}  & \cred{0.6} & \cred{4} & 2.30 & 0.0737 & 0.8658 & 28.87 \\
        \bottomrule
    \end{tabular}
    }
    \vspace{-\intextsep}
\end{table*}

%
%
%
%
%

\section{Comparison with Concurrent Work: RAS}
\label{sec:appendix_ras}

In this section, we provide a comparative analysis with the concurrent work, Region-Aware Sampler (RAS)~\cite{ras}, to clarify the fundamental conceptual differences and superior performance of \sys. Both our work and RAS address the same fundamental challenge: accelerating diffusion models by adaptively allocating computation across different regions. However, despite this shared high-level objective, our approaches are built on fundamentally different philosophies and technical foundations. 

RAS adopts a \textbf{reactive, observation-driven strategy}. It is predicated on the intuition that a model's computational focus shifts throughout the generation process. Consequently, it relies on continuous, step-by-step monitoring, employing a simple heuristic—the variance of the predicted noise in a patch—to decide which regions to update. This methodology is limited in two critical ways: first, the underlying observation of a "shifting focus" is not rigorously validated. Second, the chosen heuristic is too simplistic to accurately capture a region's true computational demand. This results in imprecise region identification and significantly inferior performance.

In stark contrast, \sys introduces a \textbf{proactive, theory-driven paradigm}. Our work is grounded in a more fundamental principle, which we extensively motivate and verify in Section~3: a region's inherent content complexity is directly correlated with its convergence requirements. \sys leverages this insight to foresee the entire computational trajectory for every region. This prediction is made possible by a theoretically robust metric derived from flow-matching principles (detailed in Section~4.2) and is accomplished in a single shot within the early stages of generation.

Therefore, the core novelty of \sys lies in its demonstration, both theoretically and empirically, that regional computational requirements can be \textbf{predicted} during early generation stages. This early-stage awareness enables \sys to precisely optimize resource allocation, resulting in superior acceleration while maintaining high generation quality across both image and video generation tasks on much newer diffusion models.

\subsection{Empirical Comparison}

\begin{figure}[h]
    \centering
    \includegraphics[width=\linewidth]{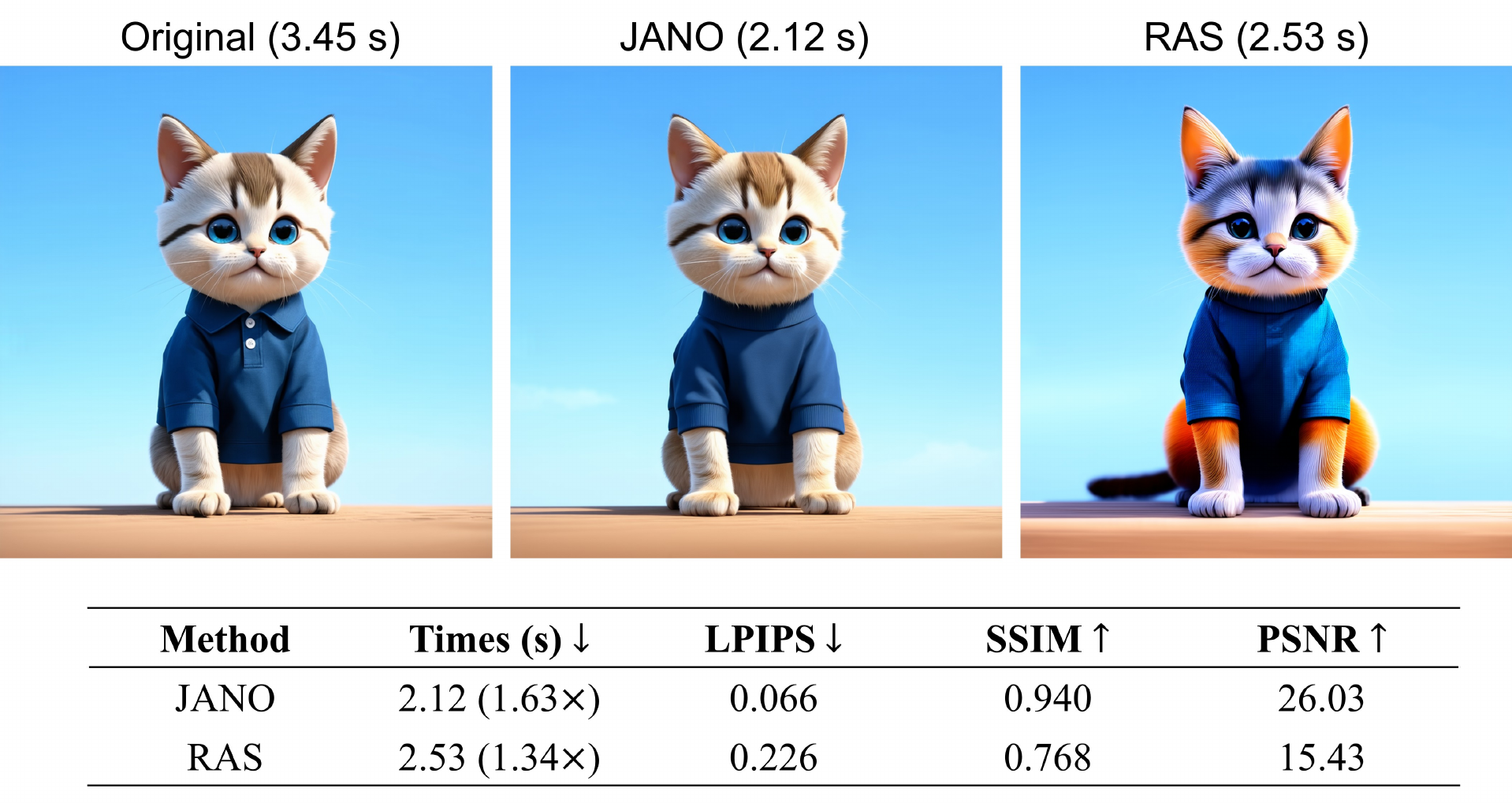}
    \caption{Performance comparison between \sys and RAS on Stable Diffusion 3, showing generated images and latency-quality analysis. Test prompt: "A photorealistic cute cat wearing a simple blue shirt, standing against a clear sky background."}
    \label{fig:visual_ras}
\end{figure}

To quantitatively validate the advantages of our approach, we implemented \sys and compared it with RAS's official implementation on Stable Diffusion 3~\cite{sd3} under identical settings. As demonstrated in Fig.~\ref{fig:visual_ras}, \sys consistently outperforms RAS in both generation speed and output quality, corroborating our theoretical analysis.

\paragraph{Region Selection Strategy}

\begin{figure}
    \centering
    \includegraphics[width=\linewidth]{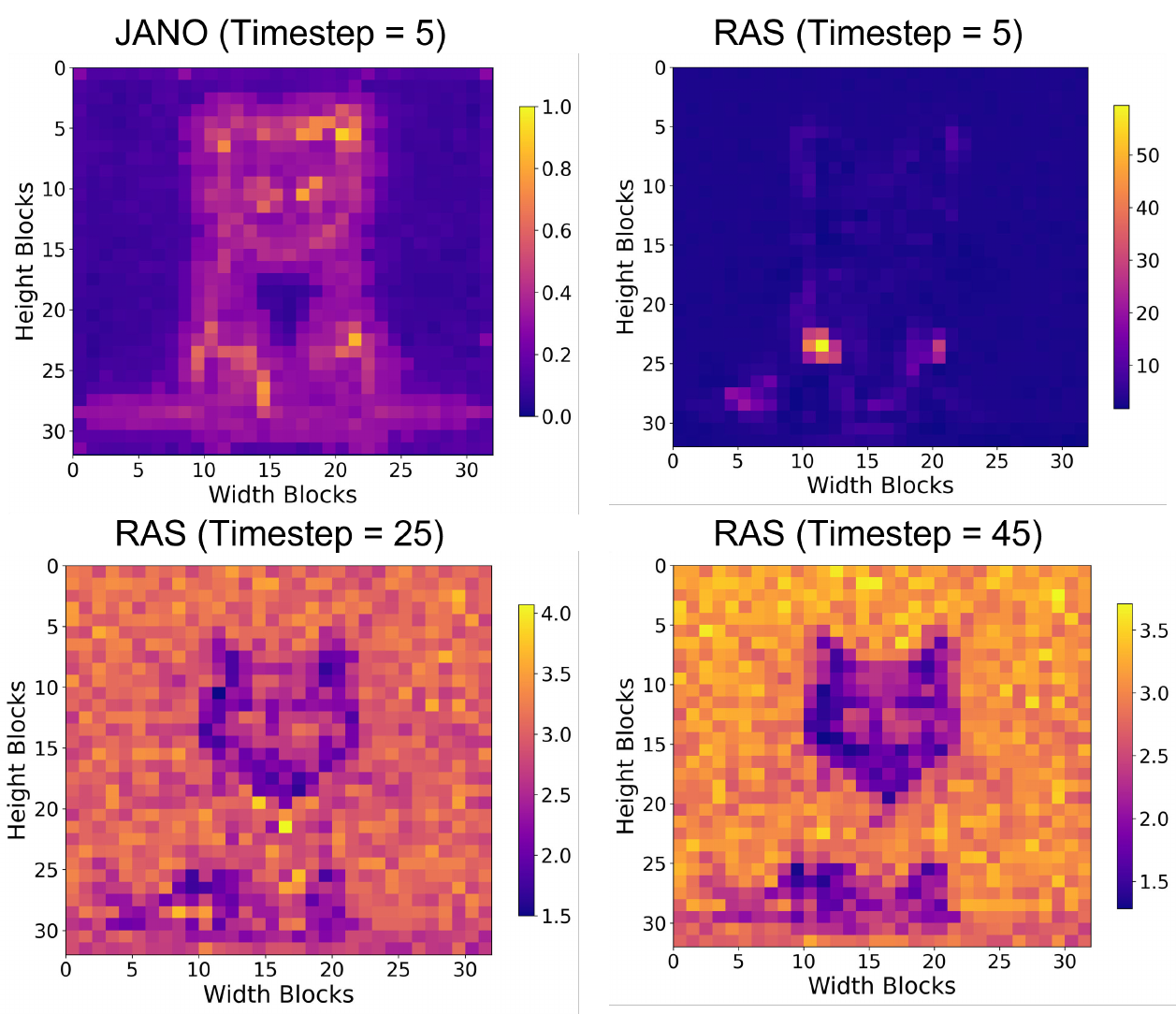}
    \caption{Comparison of JANO complexity maps and RAS token-importance maps at different diffusion timesteps.}
    \label{fig:enter-label}
\end{figure}

RAS is essentially designed to identify, at each diffusion timestep, the locally emphasized regions of generation, whereas JANO aims to infer the global image complexity already at early stages of the process. Fig.~\ref{fig:enter-label} illustrates this fundamental distinction: for RAS, regions with low prediction variance correspond to the tokens the model is currently focusing on, but at timestep 5 it still fails to capture the overall image complexity, and the identified regions drift as the timestep evolves. In contrast, JANO is able to reliably estimate the global complexity of the target image as early as timestep 5.

\section{Memory Overhead}

\sys requires additional memory to store KV outputs for static and moderate tokens at each layer, introducing notable memory overhead. Table~\ref{tab:memory} shows the memory consumption before and after enabling KV cache:

\begin{table}[h]
    \centering
    \footnotesize
    \caption{Memory consumption comparison with different caching strategies.}
    \begin{tabular}{c|c|c}
        \hline
        \textbf{Model} & \textbf{Base Memory} & \textbf{With KV Cache} \\
        \hline
        Flux-1 & 32.47 GB & 34.94 GB \\
        Wan-1.3B & 7.57 GB & 28.18 GB \\
        Wan-14B & 65.02 GB & 269.02 GB \\
        \hline
    \end{tabular}
    \label{tab:memory}
    \vspace{-\intextsep}
\end{table}

As model size and sequence length increase, the KV cache memory footprint becomes significant. To address this, we implement two optimization strategies: (1) reducing memory usage to 1/4 through conditional-free guidance parallelism and hidden state caching instead of KV caching for Wan-14B; (2) developing an asynchronous stream-based KV cache CPU offloading mechanism that enables single-GPU execution of Wan-14B with only 10\% additional latency compared to the non-offloaded version.

\section{Visualization of \sys}
\begin{figure*}[p]
    \centering
    \includegraphics[width=\textwidth]{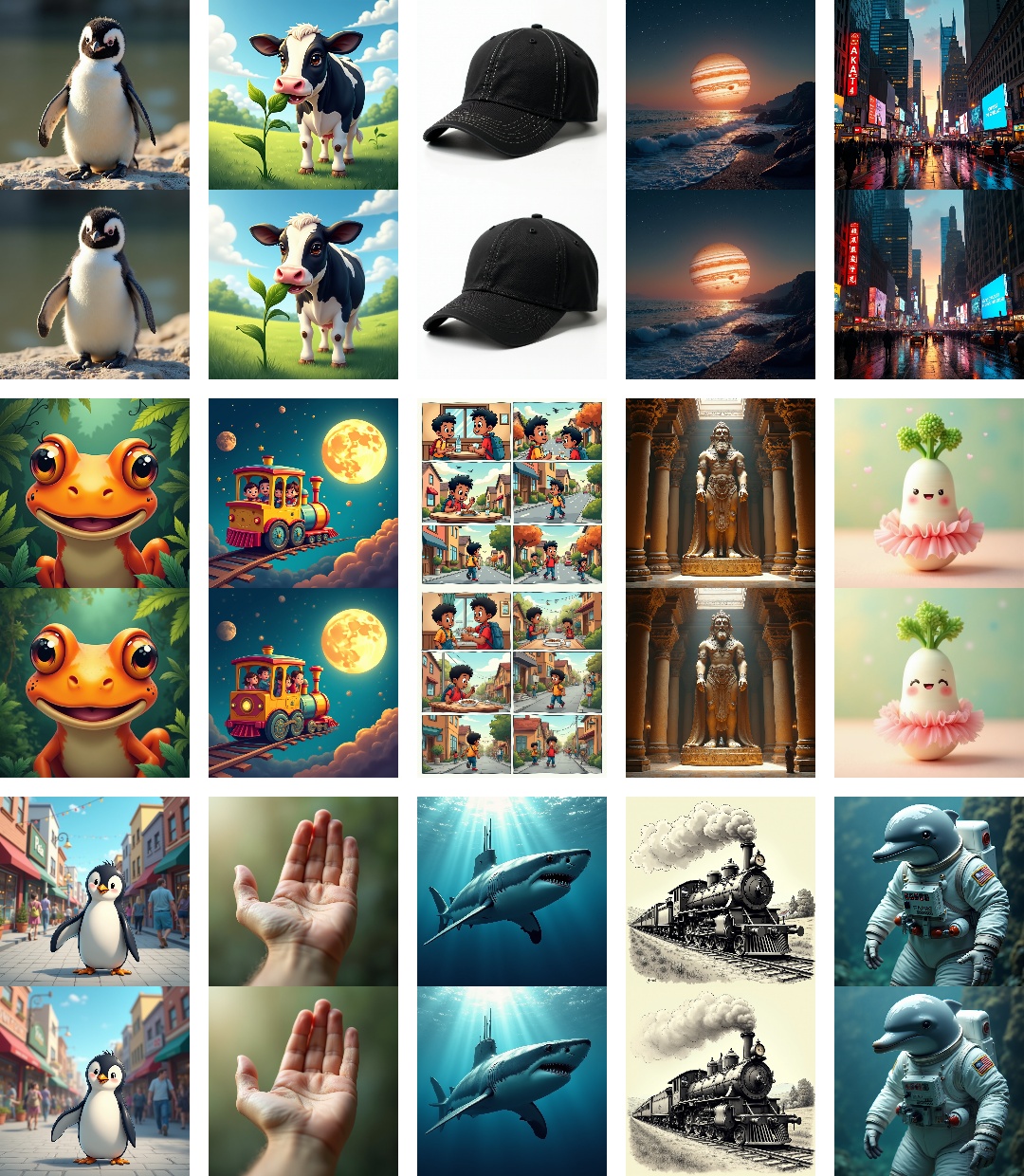}
    \caption{Flux visualization comparison between original images (top) and \sys results (bottom).}
    \label{fig:flux_visual}
\end{figure*}

\begin{figure*}
    \centering
    \includegraphics[width=\textwidth]{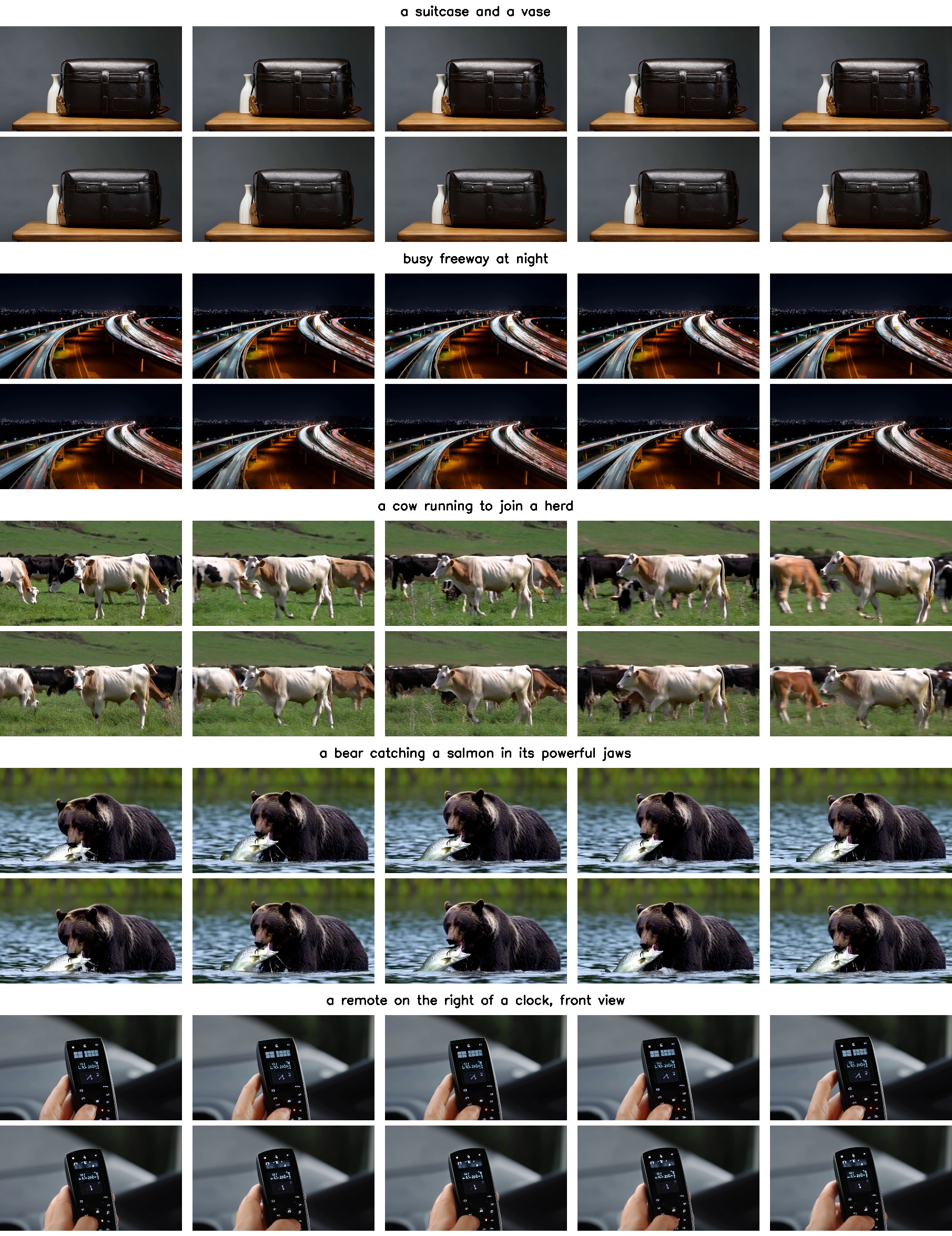}
    \caption{Wan-1.3B visualization comparison between original videos (top) and \sys results (bottom).}
    \label{fig:wan1.3B_visual}
\end{figure*}

\begin{figure*}
    \centering
    \includegraphics[width=\textwidth]{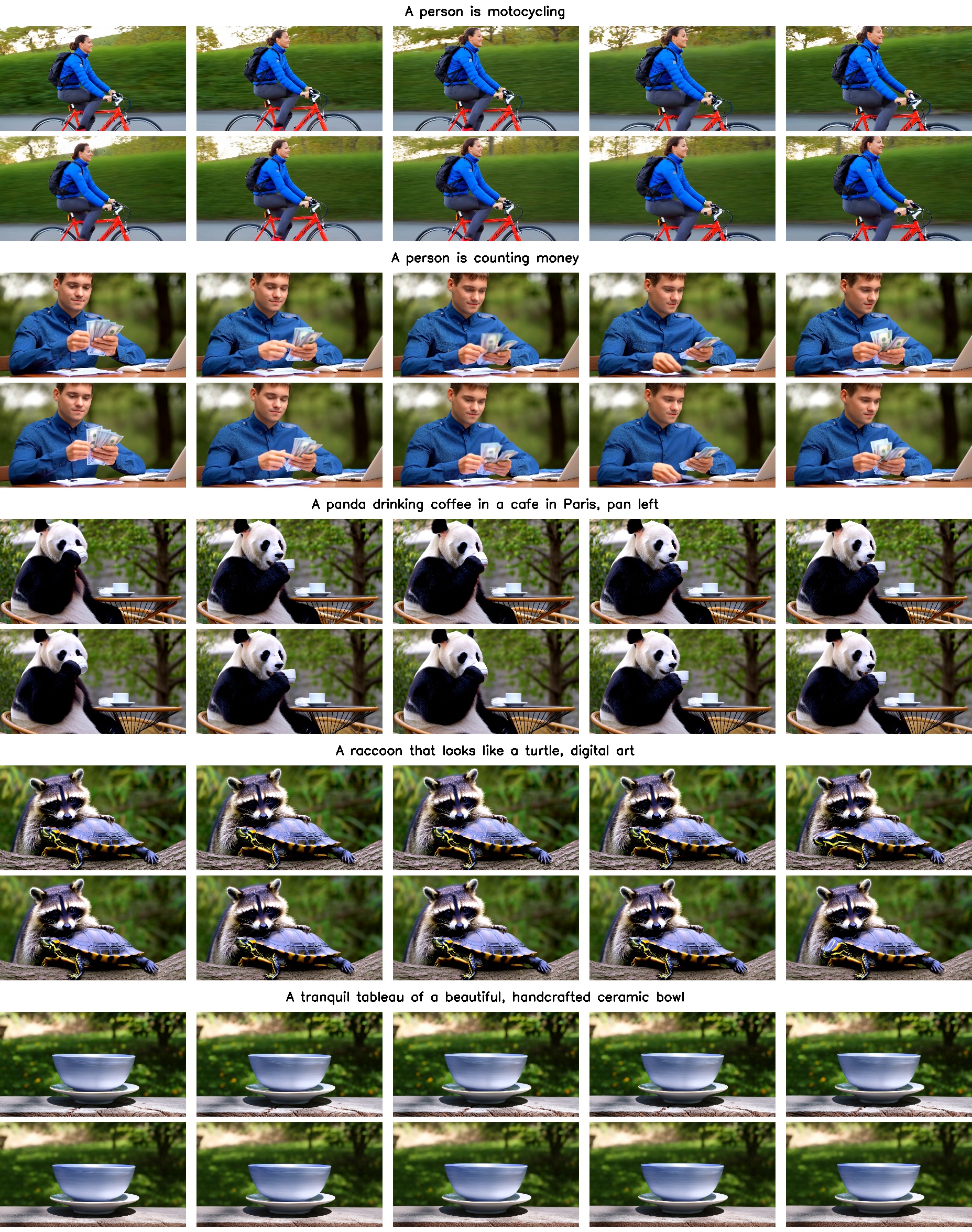}
    \caption{Wan-14B visualization comparison between original videos (top) and \sys results (bottom).}
    \label{fig:wan14B_visual}
\end{figure*}

We provide visualization comparison between original video generation and \sys on Flux-1 and Wan 2.1, as shown in Fig.~\ref{fig:flux_visual}, Fig~\ref{fig:wan1.3B_visual} and Fig.~\ref{fig:wan14B_visual}. We conduct both Text-to-Image generation and Text-to-Video generation under model-default resolution. Results demonstrates \sys can preserve high pixel-level fidelity, achieving similar generation quality compared with the original generation.

\end{document}